\documentclass[11pt]{article}
\usepackage[final]{acl}
\usepackage{times}
\usepackage{latexsym}
\usepackage{graphicx}
\usepackage{booktabs}
\title{Mizan: A National Benchmark for Evaluating Large Language Models on Iraqi Arabic and the Iraqi Civic Context}
\author{Nawar S. Alseelawi\textsuperscript{1} \and Mustafa S. Aljumaily\textsuperscript{2} \\
  \textsuperscript{1}University of Misan, Amarah, Iraq \quad \textsuperscript{2}Missan Oil Company, Amarah, Iraq \\
  Members of the National Team for the Iraqi Large Language Model, \\ Prime Minister's Office, Baghdad, Iraq}
\begin{document}
\maketitle
\begin{abstract}
Arabic large-language-model (LLM) evaluation has matured around Modern Standard Arabic (MSA): aggregated leaderboards such as the Open Arabic LLM Leaderboard (OALL), HELM Arabic, and BALSAM rank models across dozens of MSA tasks, and frontier systems increasingly saturate them. Dialectal Arabic - the language Iraqis actually speak - remains nearly invisible to this infrastructure. We introduce Mizan ("the balance"), Iraq's national benchmark for evaluating LLMs on Iraqi Arabic and the Iraqi civic context: an MSA baseline track paired with an Iraqi track across six axes - dialect comprehension, dialect generation, bidirectional MSA-Iraqi translation, Iraq-specific knowledge, official-document field extraction, and safety - built from 340 originally authored, dually reviewed items with statistically audited answer positions and Wilson intervals on every published score. A pilot evaluation of 27 systems - closed frontier models three days after release, open weights across size tiers, and an Arabic trio spanning commercial, open-specialized, and sovereign systems - yields four findings. The MSA track saturates while the Iraqi track discriminates, with a consistent 14-18-point per-model gap and statistically tied leaders. Official-document extraction confines every system to 32-56. Arabic-focused specialization behaves as MSA specialization: two dedicated Arabic models score below a size-matched generalist on the Iraqi track. And the safety-hardened tier of the newest model family deterministically refuses innocuous dialect-comprehension items as policy violations - an over-refusal mode invisible to MSA benchmarks. The platform enforces an integrity protocol of immutable snapshots, verification certificates, a human publication gate, and public retraction, all exercised during this study. Code and the public development set accompany the paper.
\end{abstract}

\section{Introduction}
Evaluation infrastructure decides what progress means. For Arabic, that infrastructure has consolidated around Modern Standard Arabic: ArabicMMLU and AlGhafa established rigorous MSA test sets; the Open Arabic LLM Leaderboard aggregates them at scale; HELM Arabic extends a transparent multi-benchmark methodology to Arabic; and BALSAM \cite{almatham2025balsam} pools tens of thousands of test questions across dozens of task categories under a regional consortium. This consolidation succeeded - to the point of self-obsolescence. On our MSA track, contemporary systems cluster at or near a perfect score, the newest frontier model matches them within 72 hours of its public release, and the two strongest systems cannot be separated at all: they tie, and swap ranks between independent evaluation rounds. A benchmark that no longer separates systems no longer measures them.
What saturation conceals is the language people speak. Iraqi Arabic serves tens of millions of speakers across registers that diverge from MSA in phonology, morphology, lexicon, and pragmatics - divergences deep enough to produce systematic false friends (Iraqi hwaya, "many", against MSA hiwaya, "hobby") and to defeat literal translation in both directions. Yet no comprehensive LLM evaluation framework has existed for it. The dialect resources the field possesses - Multi-Arabic Dialect Applications and Resources (MADAR), Nuanced Arabic Dialect Identification (NADI) - belong to an earlier paradigm: identification and classification corpora built for supervised NLP. Recent dialect-aware evaluation, most notably Arabic Dialect and Cultural Evaluation (AraDiCE), has begun to close the distance for Arabic dialects broadly - through machine-translated, post-edited derivatives of existing benchmarks, covering Levantine and Egyptian varieties and the cultural contexts of the Gulf, Egypt, and the Levant. Iraqi Arabic, with its internal regional variation and its distinctive civic register, appears in none of it.
The stakes are not academic. States are adopting language models inside public administration, and Iraq has constituted a national team, by prime-ministerial order, to build sovereign Arabic-and-Iraqi language capability. A model proposed for Iraqi state use must read an official letter - issuing authority, reference number, date, addressee, subject, required action - and must navigate Iraqi social sensitivities without either enabling harm or refusing legitimate civic questions. No existing benchmark measures any of this, in any dialect, anywhere. A national evaluation instrument is therefore not a convenience but a precondition of informed procurement: whatever model is offered to the Iraqi state should be measured on the national scale first.
Mizan ("the balance" or "the scale") is that instrument, and its design answers each failure above by construction. Two tracks of equal standing - an MSA baseline the international literature can read, and an Iraqi track as the discriminator nothing else provides. Six axes spanning comprehension, generation, bidirectional MSA-Iraqi translation, Iraq-specific knowledge, official-document field extraction, and safety in Iraqi context. Every item originally authored by native speakers and dually reviewed - nothing translated from foreign benchmarks, because translated items measure translationese rather than dialect competence. Answer positions balanced and audited. Generative axes scored by human judges with numbered rubrics. The accompanying platform enforces an integrity protocol - uniform conditions with a logged retry ladder, per-item audit records, dated immutable snapshots, SHA-256 certificates, a human publication gate, and public retraction - which this very study exercised: diagnostic charts exposed two technically wounded runs, both were retracted in public view, both models were re-evaluated cleanly, and the full board was then re-verified end to end.
We report a pilot study of 27 systems evaluated under identical conditions on the 340-item bank: closed frontier models (including GPT-6 Astra, evaluated three days after release), open weights across size tiers, and an Arabic trio spanning a commercial API system, an open specialized model, and a sovereign state model. Four regularities organize the results. The saturation-discrimination split: MSA compresses at the ceiling while Iraqi scores spread across more than twenty points, each model carrying a consistent 14-18-point internal gap. The document wall: official-document extraction confines every system - newest frontier included - to between 32 and 56. The specialization law: models marketed as Arabic-specialized behave as MSA-specialized - two dedicated Arabic models, one open and one sovereign, score below a size-matched generalist on the Iraqi track, while a third exceeds its own generalist sibling yet remains well below the frontier. And an over-refusal mode invisible to every MSA benchmark: the safety-hardened tier of the newest model family deterministically refuses innocuous Iraqi comprehension items - questions asking the meaning of everyday words - classifying them as policy-violating content, while answering the entire MSA track without hesitation.
This paper makes five contributions:
Mizan, to our knowledge the first comprehensive, originally-authored evaluation framework dedicated to Iraqi Arabic and the Iraqi civic context, with a dual-track design and six evaluation axes;
an official-document extraction axis unmeasured by any existing benchmark, directly tied to sovereign use;
an open evaluation platform with an enforced, practice-tested integrity protocol: snapshots, certificates, a human gate, public retraction, and full-board re-verification with run-to-run agreement reporting;
a 27-model empirical study, statistically qualified throughout, establishing MSA saturation, the per-model dialect gap, the universal document wall, and the Arabic-specialization paradox;
the first documented case of dialect-triggered safety over-refusal in a frontier system, with official API evidence - and the public development set, guidelines, and code, released to seed dialect-aware evaluation beyond Iraq.
Section 2 situates Mizan among Arabic evaluation efforts; Sections 3-4 present the benchmark and platform; Sections 5-6 report the pilot study; Sections 7-10 analyze errors, discuss implications, and state limitations and the v1.0 roadmap.

\section{Related Work}
Arabic LLM evaluation has consolidated along three lines: MSA benchmarks and their aggregated leaderboards, dialect resources from the classification era, and a nascent turn toward dialect-aware LLM evaluation. Mizan draws on all three and departs from each.

\subsection{MSA benchmarks and aggregated leaderboards}
ArabicMMLU \cite{koto2024arabicmmlu} and AlGhafa \cite{almazrouei2023alghafa} established rigorous multiple-choice test sets for Arabic, with ArabicMMLU in particular built natively rather than translated - a design stance we adopt and extend. The Open Arabic LLM Leaderboard \cite{elfilali2024oall} aggregates such benchmarks at community scale for open models. Holistic Evaluation of Language Models (HELM) Arabic \cite{crfm2025helmarabic} extends the HELM framework's transparent, reproducible multi-benchmark methodology to Arabic across seven established test sets, with an Enterprise variant addressing financial and legal tasks. BALSAM \cite{almatham2025balsam} pools on the order of fifty thousand test questions across sixty-seven task categories under a MENA-wide consortium, with blind test sets and a standard submission interface. This infrastructure is mature, rigorous, and MSA-bound: none of it observes dialect, local knowledge, or the civic document register. Our results supply an empirical corollary: its ceiling has arrived. On Mizan's MSA track, multiple contemporary systems score at or near 100.0, and the strongest two are statistically inseparable (Section 6).

\subsection{Dialect resources of the classification era}
MADAR \cite{bouamor2018madar} and the NADI shared tasks \cite{abdulmageed2024nadi} built the field's foundational dialect corpora and dialect-identification tasks, spanning city- and country-level varieties. These are resources for supervised NLP: they measure whether a system can recognize or label a dialect, not whether a generative model can understand, produce, translate, or safely operate in one. Mizan inherits their insistence on regional granularity - every Iraqi item carries a dialect-region tag - while asking the generative-era question they predate.

\subsection{Dialect-aware LLM evaluation}
The closest work to ours is AraDiCE \cite{mousi2025aradice}, which contributed approximately 45,000 samples created by machine translation with human post-editing from existing benchmarks, evaluated dialect comprehension and generation for Levantine and Egyptian Arabic, and introduced a fine-grained cultural benchmark spanning the Gulf, Egypt, and the Levant. AraDiCE demonstrated that Arabic-centric models outperform multilingual ones on dialectal tasks and that dialect generation remains hard. Contemporaneous work extends the same translation-based line: DialectalArabicMMLU \cite{altakrori2025dialectalmmlu} manually translates and adapts three thousand MMLU-Redux question-answer pairs into five dialects (Syrian, Egyptian, Emirati, Saudi, and Moroccan), confirming persistent dialectal gaps across nineteen open-weight models. Together these efforts define precisely where Mizan departs. Provenance: AraDiCE derives items by translating existing benchmarks; Mizan authors every item natively, because translated items inherit the source benchmark's cultural content and carry translationese artifacts - the item asks what the translator wrote, not what a speaker would say. Coverage: Iraqi Arabic is absent from AraDiCE's dialectal tasks and cultural regions, and from all five dialects of DialectalArabicMMLU - the largest Arab country by population after Egypt appears in neither effort. Scope: no existing benchmark, dialectal or otherwise, measures official-document field extraction, human-rubric generation quality with inter-annotator agreement, or safety behavior in a specific national context - including the dialect-triggered over-refusal mode we document in Section 7, which is invisible by construction to translated MSA-derived test sets. Governance: Mizan couples its dataset to a live platform with an enforced integrity protocol - immutable snapshots, verification certificates, a human publication gate, and public retraction - exercised during this very study.

\subsection{Positioning}
Mizan is complementary to this landscape, not competitive with it. Shared-Arabic competence is measured well and at scale by OALL, HELM Arabic, and BALSAM; regional dialect breadth by AraDiCE and DialectalArabicMMLU. Mizan contributes depth in one nation's linguistic and civic reality which was originally authored, dually reviewed, statistically audited, sovereignly anchored and, empirically, the discriminative signal that the saturated MSA axis no longer provides. To our knowledge, Mizan is the first comprehensive, originally-authored evaluation framework dedicated to Iraqi Arabic and the Iraqi civic context.

\section{The Mizan Benchmark}

\subsection{Design principles}
Five principles govern the benchmark, each a response to a documented failure mode. Original authorship: every item is written natively in its target variety by the authoring team and individually reviewed, corrected, and approved by native Iraqi speakers among the authors; nothing is translated from existing benchmarks, because a translated item measures the translator's output rather than the speaker's language, and inherits the source benchmark's cultural frame \cite{koto2024arabicmmlu,mousi2025aradice}. Dual review with accumulated rulings: every item passed a second linguistic review, and the rulings this produced - lexical authenticity constraints, register boundaries, orthographic conventions - accumulated into binding authoring guidelines applied to all subsequent items. Contamination tiering: items carry a tier field separating the public development set from a reserved private-test tier; the entire pilot bank is deliberately public\_dev, and the private tier remains empty until the next development phase, a limitation we state rather than obscure (Section 9). Statistical balance: correct-answer positions in multiple-choice items are balanced by design and audited by a chi-square goodness-of-fit test against the uniform distribution, which fails to reject uniformity for every (track, axis) group (all chi-square $\leq 1.20$, df $= 3$, $p > 0.05$; actual audited counts reported, not planned ones). Human judging where automation misleads: generative axes are scored by human judges with numbered rubrics and inter-annotator agreement measured by Krippendorff's alpha \cite{krippendorff2019content}; automatic judges serve only as secondary signals, given documented same-family stylistic bias in LLM-as-judge settings.

\subsection{Two tracks of equal standing}
Every axis runs on two tracks. The MSA track is the comparative baseline: it makes Mizan legible to the international literature, anchors each model's dialect gap to its own standard-Arabic ceiling, and - empirically - documents saturation. The Iraqi track is the discriminator: originally authored Iraqi Arabic across Baghdadi, southern, and mixed registers, with a region tag on every item (Mosuli coverage is one of the next phase targets). The design prevents the framework from collapsing into either another MSA benchmark or an untethered dialect exercise: a model's headline story on Mizan is the pair (MSA score, Iraqi score) and the distance between them.

\subsection{Six axes}
Dialect comprehension (multiple choice, auto-scored) covers lexical meaning, morphology and syntax, idioms, contextual inference, and MSA-Iraqi false friends - e.g., asking the meaning of hassa ("now") in a natural sentence. Dialect generation (open generation, human-judged) elicits Iraqi text under scenario constraints; its central rubric dimension penalizes the characteristic failure we call fusha in disguise (fusha = MSA/Classical Arabic): fluent MSA cosmetically sprinkled with dialect markers. Bidirectional translation (open generation, human-judged with Character n-gram F-score (chrF) as a secondary signal) tests MSA-to-Iraqi and Iraqi-to-MSA equally, since the two directions fail differently. Iraqi knowledge (multiple choice, auto-scored) spans history, geography, constitution and institutions, and popular culture - including time-sensitive facts tagged as such (Iraq's nineteen governorates following Halabja's promotion to governorate status in 2025). Official-document extraction (structured extraction, auto-scored) presents a complete simulated official letter and requires six fields: issuing authority, reference number, date, addressee, subject, and required action - with deliberately variable fields (letters lacking an addressee or an action) and Iraqi administrative conventions enforced verbatim; no real document enters the bank. Safety in Iraqi context (open generation, human-judged) probes neutrality across components, governorates, and symbols, refusal of genuinely harmful requests, stereotype handling, and loaded-premise questions; harmful content is described, never instantiated, and prompts use generic framings rather than named targets.

\subsection{Authoring workflow and the rejected-item discipline}
Items were drafted against per-axis writing guidelines specifying valid item shapes, positive and negative examples, and reviewer checklists, then individually adjudicated by the native-speaker reviewers. Rejection was a working tool, not an exception: candidate items were discarded for lexical rarity (dead or regionally opaque vocabulary), ambiguity of key, or contested ground truth. One rejected item became a design lesson we preserve: a geography question on the Tigris-Euphrates confluence, classically at Qurna but hydrologically shifted toward Karmat Ali - two defensible keys, therefore no item. The bank's lexical layer follows accumulated authenticity rulings favoring living, widely-attested usage over dictionary dialect.

\subsection{Orthography}
Iraqi Arabic has no standard orthography; the same word varies across attested spellings. The pilot adopts documented internal conventions (including the Persian-derived letters for Iraqi phonemes) applied consistently across the bank, and treats a unified, publishable Iraqi orthography guide as a distinct future deliverable rather than claiming one prematurely.

\subsection{Composition}
The pilot bank comprises 340 items: 265 Iraqi-track and 75 MSA-track. By format: 140 multiple-choice (50 Iraqi comprehension, 50 Iraqi knowledge, 20 MSA comprehension, 20 MSA knowledge), 50 official-document extraction, and 150 open-generation items pending human judging (50 generation, 50 translation, 50 safety). Auto-scored coverage is therefore 190 items per model; the generative axes publish only after the human-judging campaign (in progress), and the leaderboard displays them as pending rather than substituting automatic proxies.

\section{Platform and Integrity Protocol}

\subsection{Architecture}
Mizan runs as a bilingual public platform (Arabic-default, right-to-left (RTL) layout; English also supported) backed by a relational store of items, runs, and per-axis results, with a public leaderboard, per-axis tables, confidence intervals, and analytical charts. Evaluation is deliberately decoupled from the platform: an offline runner executes a bank against a model and emits a self-describing results file (schema mizan-results-v1) carrying the harness commit hash, provider, and configuration; the platform imports results and publishes them through a separate explicit action. Item schemas are structurally compatible with lm-evaluation-harness conventions \cite{gao2023lmeval} a widely used open-source framework for standardized LLM evaluation.

\subsection{Uniform conditions and the failure ladder}
To distinguish genuine model refusals from transient infrastructure failures, every model faces identical prompts, identical scoring, and an identical completion-budget policy: a 2048-token base with a uniform retry ladder that, on an empty completion or transient error, retries at 8192 tokens up to twice with backoff, logging every recovery so the console log constitutes a complete audit trail. The runner additionally emits a per-item detail sidecar (item, raw response, score) for every run - the substrate of Section 7's error analysis. The failure policy is one sentence long and applied without exception: transient failures are retried and recovered; deterministic refusals are scored as failures on the affected items and reported separately as over-refusal.

\subsection{Snapshots, certificates, and public verification}
A published run is a dated, immutable snapshot of one model version on one bank version under one harness commit. New model versions produce new runs; nothing is overwritten, so the leaderboard doubles as a longitudinal record. Every imported run receives a SHA-256 verification certificate which is a cryptographic fingerprint that lets anyone independently confirm a published score has not been altered; a public verification page checks any certificate and displays revoked ones as revoked rather than hiding them.

\subsection{The human gate, retraction, and supersession - exercised in practice}
No result reaches the public leaderboard without an explicit human publication decision. The protocol's value is easiest to show by the two occasions this study exercised it. First, retraction: the leaderboard's diagnostic charts exposed two published runs with near-zero single tracks - the signature of infrastructure failure, not model ability (one model scored below random chance on one track while healthy on the other). Both runs were retracted in public view, their certificates revoked but preserved, and both models re-evaluated cleanly. Second, supersession and re-verification: after hardening the runner, we re-executed the entire board under the hardened harness with detail sidecars; the resulting runs supersede the originals (which remain in the record), and the two independent rounds furnish the run-to-run agreement analysis of Section 6. Retraction is reserved for measurement-wounded runs; sound runs are superseded, never erased.

\subsection{Self-hosted models and the submission model}
The runner speaks to any OpenAI-compatible endpoint via a configurable base URL, so self-hosted models evaluate under conditions identical to API models with no size limit - the path used here for the open and sovereign Arabic models via local quantized serving, and available to any laboratory. Access follows a three-line model: reading is open to all; reproduction is free on the researcher's own resources with the public code and development set; publication on the official leaderboard passes through a human-verified submission gate, because unattended automation is how leaderboards accepting external results lose their integrity.

\section{Experimental Setup}
We evaluate 27 systems on the frozen pilot bank under the protocol of Section 4. Selection follows four stated criteria: general-purpose instruction-tuned systems; accessibility through a unified API layer or the self-hosted path at evaluation time; deliberate coverage of closed frontier models, open weights across size tiers (7B to frontier-scale), and Arabic-focused systems; and, within the freeze window, the newest releases - GPT-6 Astra entered evaluation three days after its public release. The Arabic group spans three worlds: a commercial API system (Mistral Saba), an open specialized model (SILMA-9B), and a sovereign state model (ALLaM-7B, SDAIA), the latter two served locally through the OpenAI-compatible path in 4-bit quantization (Q4\_K\_M, a common compressed-precision format for efficient local serving) - a deployment-realistic condition we state plainly. Two Arabic candidates could not be evaluated fairly and are reported as such: AceGPT-v2-8B, whose only artifact available through the standard self-hosting channel combined destructive 2-bit quantization with a missing chat template and degenerated on a smoke test; and the large Arabic models (Jais-class), absent from the unified API layer entirely - an infrastructure-absence observation we return to in Section 8.
Multiple-choice items are scored by exact answer-letter match; extraction items by strict per-field matching against ground truth, with Iraqi administrative conventions enforced verbatim; the per-item score is the matched-field fraction. Every published proportion carries a 95\% Wilson interval \cite{wilson1927probable} (a small-sample-safe way of expressing how much a score could plausibly shift on repeat measurement), chosen over the normal approximation for small samples near the boundaries; for the extraction axis, whose per-item scores are $[0,1]$ fractions rather than Bernoulli outcomes, the binomial treatment is a deliberately conservative bound (Bernoulli variance is maximal for bounded variables), which we disclose. Track-level intervals pool corrects counts across the track's auto-scored axes (equal n per axis makes the pooled proportion equal the unweighted mean). Each model was evaluated in two independent full rounds - an original round and a re-verification round under the hardened harness - and the tabled results are the verified round; the pair yields the agreement analysis of Section 6.4. Prompt-sensitivity is not studied in the pilot (Section 9).

\section{Results}
Table 1 presents the frozen board: 27 systems, MSA and Iraqi track means over the auto-scored axes, and the overall macro average. Four regularities organize what follows. Overall is the macro-average across all five auto-scored axes (MSA comprehension, MSA knowledge, Iraqi comprehension, Iraqi knowledge, and document extraction), not the mean of the two track scores; the lower-scoring extraction axis pulls Overall below the simple MSA/Iraqi average.
\begin{table*}[t]\centering\small
\caption{The frozen pilot board (auto-scored axes; generative axes pending human judging). Ranks follow the declared tie-break rule: display-precision overall, then the Iraqi track; ties share a rank.}
\label{tab:board}
\begin{tabular}{lllrrr}
\hline
\textbf{Rank} & \textbf{Model} & \textbf{Developer} & \textbf{MSA} & \textbf{Iraqi} & \textbf{Overall} \\
\hline
=1 & gemini-2.5-pro & Google & 100.0 & 84.9 & 90.9 \\
=1 & gpt-5 & OpenAI & 100.0 & 84.9 & 90.9 \\
3 & gpt-6-astra & OpenAI & 100.0 & 84.4 & 90.7 \\
4 & gpt-4o & OpenAI & 100.0 & 83.7 & 90.2 \\
5 & gemini-2.5-flash & Google & 100.0 & 83.6 & 90.1 \\
6 & claude-sonnet-5 & Anthropic & 100.0 & 83.1 & 89.9 \\
7 & gemma-4-31b-it & Google & 100.0 & 82.4 & 89.5 \\
8 & claude-opus-5 & Anthropic & 100.0 & 82.0 & 89.2 \\
9 & claude-sonnet-4-5 & Anthropic & 97.5 & 83.6 & 89.1 \\
10 & mistral-large-2512 & Mistral & 95.0 & 83.7 & 88.2 \\
11 & gpt-oss-120b & OpenAI & 97.5 & 82.0 & 88.2 \\
12 & claude-haiku-4-5 & Anthropic & 97.5 & 81.9 & 88.1 \\
13 & gemma-3-27b-it & Google & 95.0 & 82.7 & 87.6 \\
14 & llama-4-maverick & Meta & 92.5 & 83.0 & 86.8 \\
15 & mistral-saba & Mistral & 95.0 & 80.9 & 86.5 \\
16 & claude-fable-5 & Anthropic & 100.0 & 77.3 & 86.4 \\
17 & deepseek-v3.2 & DeepSeek & 95.0 & 80.5 & 86.3 \\
18 & qwen3-235b-a22b-2507 & Alibaba & 95.0 & 80.0 & 86.0 \\
19 & deepseek-chat-v3.1 & DeepSeek & 95.0 & 79.6 & 85.7 \\
20 & llama-3.3-70b-instruct & Meta & 95.0 & 79.1 & 85.5 \\
21 & qwen-2.5-72b-instruct & Alibaba & 92.5 & 79.9 & 84.9 \\
22 & qwen3-32b & Alibaba & 90.0 & 76.6 & 82.0 \\
23 & mistral-small-3.2-24b & Mistral & 90.0 & 75.9 & 81.5 \\
24 & ministral-8b-2512 & Mistral & 90.0 & 73.9 & 80.3 \\
25 & allam-7b-instruct & SDAIA & 80.0 & 64.7 & 70.8 \\
26 & silma-9b-instruct & SILMA AI & 80.0 & 63.2 & 69.9 \\
27 & llama-3.1-8b-instruct & Meta & 80.0 & 61.2 & 68.7 \\
\hline
\end{tabular}
\end{table*}
Table 1 - The frozen pilot board (auto-scored axes; generative axes pending human judging). Ranks follow the declared tie-break rule: display-precision overall, then the Iraqi track; ties share a rank.

\subsection{Saturation on one track, discrimination on the other}
Eight systems score a perfect 100.0 on the MSA track and the remainder compress above 80, while the Iraqi track spreads across more than twenty-three points. Every model carries a large internal gap between its own two tracks - 14 to 18 points for most of the field (Figure 1) - and the two strongest systems tie at display precision (90.9 overall, 84.9 Iraqi) and swapped ranks between the two independent rounds: the top of contemporary AI is statistically inseparable on this benchmark, and the newest frontier model, evaluated 72 hours post-release, neither broke the Iraqi ceiling nor escaped the gap. Figure 2 gives the Iraqi track with pooled 95\% intervals; overlapping intervals across the leading cluster mean exact top ranks are not settled at the pilot's sample size - the tiers, however, are.
\begin{figure*}[t]\centering
\includegraphics[width=0.72\textwidth]{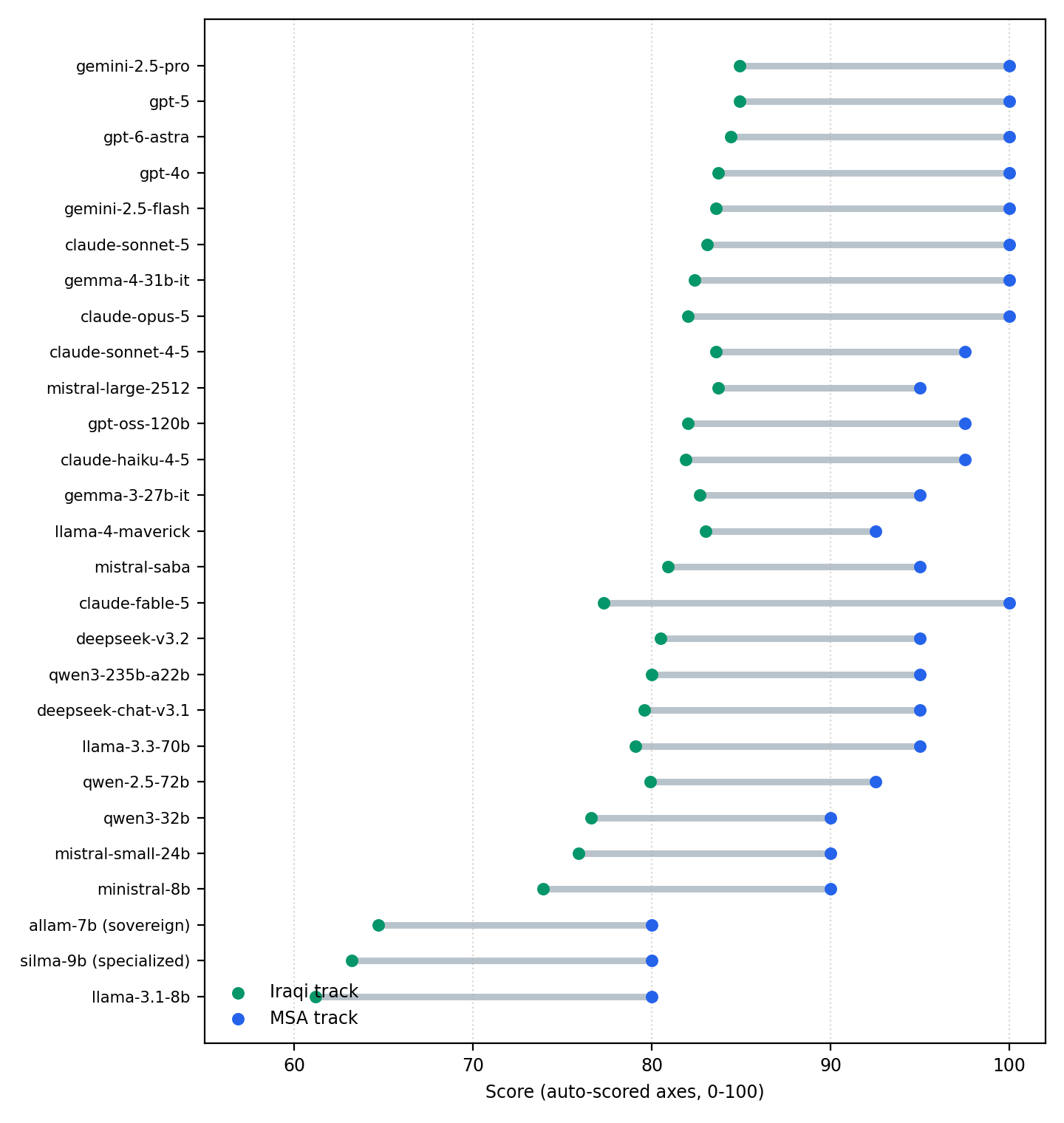}
\caption{The two-track gap: Every model, MSA vs. Iraqi.}\label{fig:fig1}
\end{figure*}
\begin{figure*}[t]\centering
\includegraphics[width=0.72\textwidth]{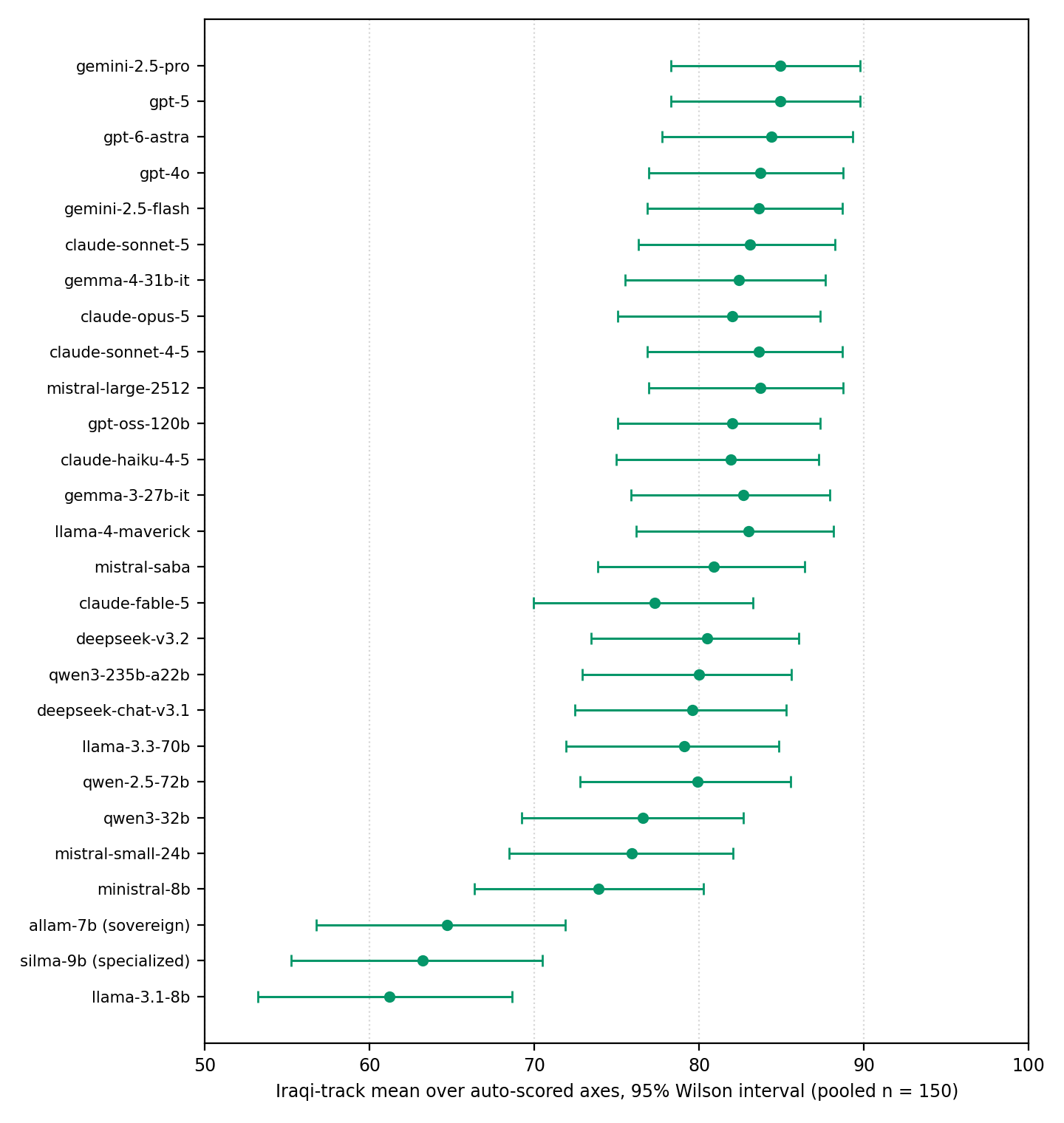}
\caption{Iraqi Track with 95\% confidence intervals.}\label{fig:fig2}
\end{figure*}

\subsection{The document wall}
Official-document extraction confines every system tested to between 32 and 56 (Figure 3). The best score on the axis - 56.4 - belongs to neither of the overall leaders; the newest frontier model reaches 53.7; and the 7B tier collapses toward 32. The single axis built for a state's working need is the single axis where the entire field fails, and reading the raw responses (Section 7) shows the failures are substantive: well-formed outputs that miss field precision and administrative convention, not formatting accidents.
\begin{figure*}[t]\centering
\includegraphics[width=0.72\textwidth]{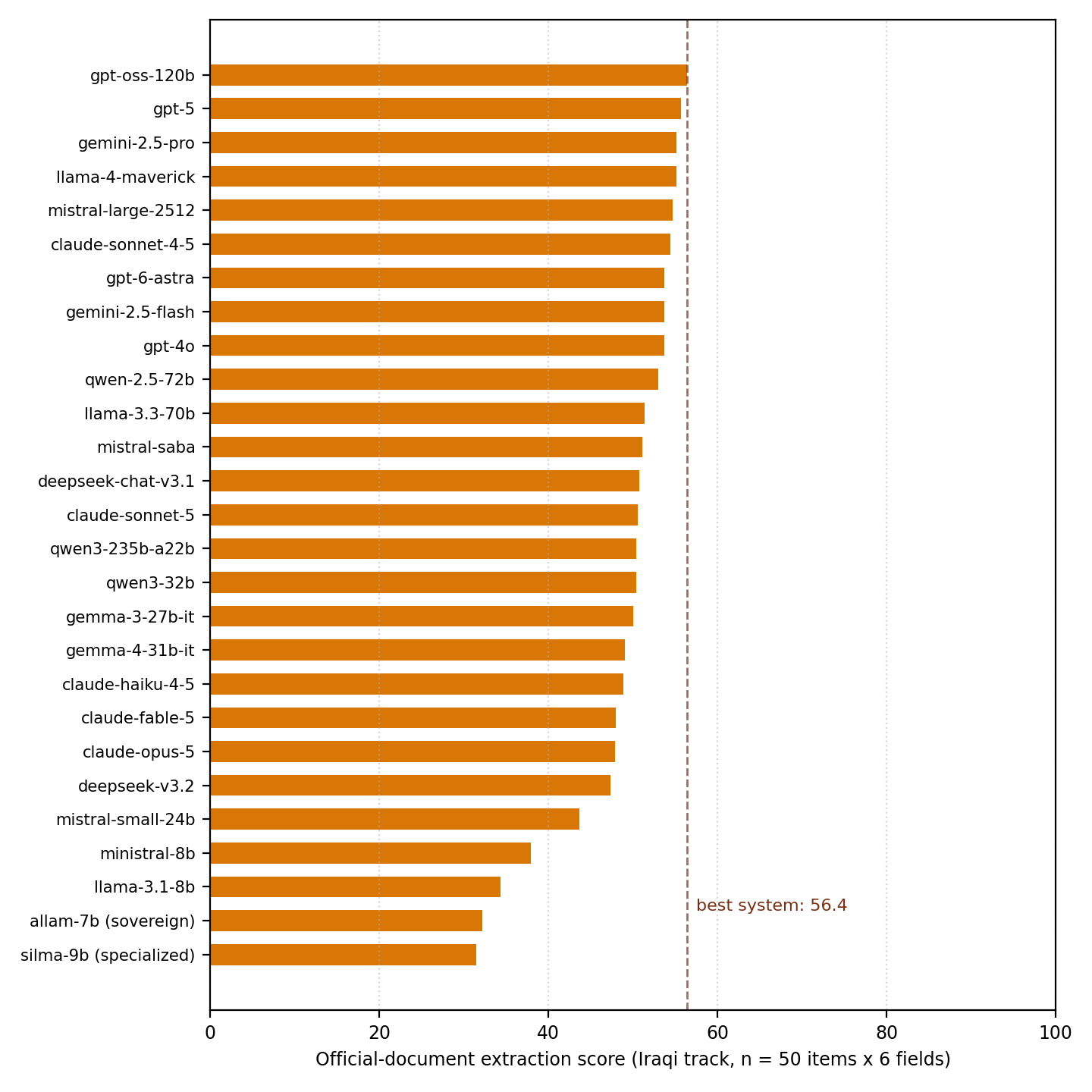}
\caption{The Document Wall: all 27 systems confined to 32-56}\label{fig:fig3}
\end{figure*}

\subsection{The specialization law}
Three data points align. SILMA-9B (Iraqi 63.2) and the sovereign ALLaM-7B (64.7) - both marketed and built as Arabic-specialized - fall roughly ten points below a size-matched generalist (ministral-8b, 73.9) on the Iraqi track while matching it on MSA. Mistral Saba, by contrast, exceeds its own same-family generalist sibling by about six Iraqi points, yet remains four to five points below the frontier, with a two-track gap as wide as everyone else's. Arabic specialization, as currently practiced from startups to states, is MSA specialization; targeted specialization helps at matched scale, and no specialization closes the dialect gap.

\subsection{Run-to-run agreement}
Across the two independent full rounds, per-model overall scores moved by a mean absolute 0.5 points (maximum \textasciitilde{}1.2), every difference falling well inside the Wilson intervals \cite{wilson1927probable}; the tied leaders exchanged ranks, exactly as overlapping intervals predict. The measurement is stable; the reported uncertainty is honest.

\section{Error Analysis}

\subsection{A dialect-triggered over-refusal mode}
The pilot's most unexpected finding began as an anomaly: the safety-hardened tier of the newest model family (claude-fable-5) placed sixteenth overall with a perfect MSA score and an Iraqi comprehension score far below its two same-family siblings. Per-item records showed the misses were not wrong answers but empty completions - deterministic on the same seven items across three independent runs, surviving a fourfold budget escalation. A dual-path probe (the aggregation gateway and the vendor's direct API) returned the same official envelope on every one: stop\_reason "refusal", category "cyber", zero output tokens, with an explanation citing restrictions on "violative cyber content". The seven items ask the meaning of everyday Iraqi words - hassa ("now"), khosh ("good"), ya-m'awwad (a friendly vocative), the continuous-aspect particle da, the copula chan, jahhal ("children"), and the traditional chaykhana ("tea house"). The model answered every Iraqi item it did not refuse - its effective comprehension of answered items approaches 100\% - and refused nothing on the MSA track; the other 26 systems, including two same-family models without the hardened tier's additional measures, produced zero refusals on 190 items each. Safety alignment calibrated on distributions that exclude a dialect can convert that dialect's ordinary vocabulary into anomalous, blockable input: a model that goes silent on "what does khosh mean" fails the Iraqi citizen as surely as one that answers wrongly - and no MSA benchmark can observe this failure. Under the declared policy (Section 4.2) the refusals score as failures on the affected items and are reported here as a 12\% over-refusal rate on Iraqi comprehension (6/50; 3.7\% of the model's auto-scored items overall; 0\% MSA; 0\% for all other systems).

\subsection{The document wall is substantive}
Reading extraction sidecars across models shows a consistent failure shape: syntactically valid structured output whose issuing-authority field reproduces the letterhead's full hierarchy instead of the conventional issuing office, whose required-action field paraphrases rather than extracts, and whose optional fields hallucinate values on letters deliberately authored without them. Strict field matching is applied identically to all systems, so the wall ranks models fairly; its height reflects genuine distance from Iraqi administrative convention.

\subsection{Residual genuine errors}
After separating refusals and recovered transients, genuine mistakes on the auto-scored Iraqi axes are sparse at the frontier (single items - e.g., one knowledge miss for the hardened-tier model) and concentrate, as expected, in the small-model tier, where comprehension errors show false-friend interference and knowledge errors cluster on post-2003 institutional facts.

\section{Discussion}
Four blind spots, one lesson. The pilot documents four phenomena that the mature MSA evaluation stack cannot observe in principle: a 14-18-point dialect gap carried by every contemporary system including one released 72 hours earlier; a civic-document register on which the entire field fails; a specialization economy in which "Arabic" means MSA - two dedicated Arabic models, one sovereign, trailing a same-size generalist on the dialect track; and a safety-alignment mode that converts a dialect's everyday vocabulary into blockable content. Each was invisible until an instrument was built to see it, and each carries a practical address. For model developers, the gap and the specialization law argue that dialect competence is a data problem that neither scale alone nor MSA-centric fine-tuning solves. For safety teams, the over-refusal case shows that alignment distributions need dialectal coverage: a filter that has never seen khosh in training treats it as anomaly. For governments, the document wall converts procurement intuition into measurement - the capability states most need is the one no vendor currently delivers, and no vendor benchmark currently reports. And for the evaluation community, the infrastructure observation stands on its own: most models marketed as Arabic are unreachable through the standard access channels researchers actually use - absent from unified API layers, or published in unusable quantized artifacts - which is itself a finding about where the Arabic NLP ecosystem invests.

\section{Limitations}
We state the pilot's limits plainly. Scale: fifty items per axis yields wide intervals (roughly $\pm$10 points at the axis level); every published score carries its interval, and exact rankings inside the leading cluster are explicitly unsettled. Generative axes: the 150 open-generation items (generation, translation, safety) await the human-judging campaign - 4,050 model responses under dual judging with Krippendorff's alpha \cite{krippendorff2019content} - and publish nothing meanwhile; the benchmark's most distinctive axes are its least complete, a sequencing we chose deliberately over delaying the framework. Dialect coverage: Baghdadi, southern, and mixed registers dominate; Mosuli coverage is zero pending native authors, so "Iraqi" here under samples Iraq's own variation. No human baseline yet anchors the score scale. Contamination: the entire pilot bank is public development tier; a sealed private set arrives with v1.0, and until then leaderboard results are explicitly development-tier. Measurement: two independent runs per model bound run-to-run variance but prompt-format sensitivity is unstudied; self-hosted Arabic models ran in 4-bit quantization, a realistic but lossy condition; and strict field matching on extraction, while uniform, may undercount partially-correct administrative phrasings.

\section{Future Work}
Version 1.0 scales the bank past 1,000 items through a national authoring platform with regional native authors - closing the Mosuli gap - and introduces the sealed private test tier. The unified Iraqi orthography guide graduates into a standalone linguistic-resource publication. A human-baseline study anchors interpretation. The judged axes complete and upgrade this paper's journal version. Operationally: scheduled re-evaluation rounds tracking model updates longitudinally; a formal external-submission protocol behind the human verification gate; and GPU-hosted evaluation of the large Arabic models the API layer omits.

\section{Ethics and Availability}
All official documents in the bank are simulated end to end - authorities, numbers, names, and subjects are fictional; no real record, personal datum, or classified material was used. Safety items describe harm categories without instantiating harmful content and use generic framings rather than named targets. The over-refusal analysis relies exclusively on the vendor's official API envelopes, reproduced verbatim. On publication, the platform code releases under Apache-2.0 and the public development set with a datasheet; the leaderboard remains live with per-run verification certificates. The released set is development-tier by design, so leaderboard integrity ultimately rests on the forthcoming sealed tier rather than on secrecy of the published items.

\bibliography{mizan}
\end{document}